\def\year{2020}\relax
\documentclass[letterpaper]{article} 
\usepackage{aaai20}  
\usepackage{times}  
\usepackage{helvet} 
\usepackage{courier}  
\usepackage[hyphens]{url}  
\usepackage{graphicx} 
\usepackage{graphicx}  
\usepackage{multirow}
\usepackage{amsmath}
\usepackage{mathbbol}
\usepackage[table,xcdraw]{xcolor}
\usepackage{CJKutf8}

\title{Creating Auxiliary Representations from Charge Definitions \\ for Criminal Charge Prediction}

\author{Liangyi Kang\textsuperscript{\rm 1}, Jie Liu\textsuperscript{\rm 1}, Lingqiao Liu\textsuperscript{\rm 2}, Qinfeng Shi\textsuperscript{\rm 2}, and Dan Ye\textsuperscript{\rm 1}\\ 
	\textsuperscript{\rm 1} Institute of Software, Chinese Academy of Sciences, Beijing, China\\ 
	\textsuperscript{\rm 2} School of Computer Science, The University of Adelaide, Australia\\ 
	\{kangliangyi15, ljie, yedan\}@otcaix.iscas.ac.cn, \{lingqiao.liu, javen.shi\}@adelaide.edu.au 
}

\begin{document}

\maketitle

\begin{abstract}
Charge prediction, determining charges for criminal cases by analyzing the textual fact descriptions, is a promising technology in legal assistant systems. In practice, the fact descriptions could exhibit a significant intra-class variation due to factors like non-normative use of language, which makes the prediction task very challenging, especially for charge classes with too few samples to cover the expression variation. In this work, we explore to use the charge definitions from criminal law to alleviate this issue. The key idea is that the expressions in a fact description should have corresponding formal terms in charge definitions, and those terms are shared across classes and could account for the diversity in the fact descriptions. Thus, we propose to create auxiliary fact representations from charge definitions to augment fact descriptions representation. The generated auxiliary representations are created through the interaction of fact description with the relevant charge definitions and terms in those definitions by integrated sentence- and word-level attention scheme. Experimental results on two datasets show that our model achieves significant improvement than baselines, especially for classes with few samples.
\end{abstract}

\section{Introduction}
The task of charge prediction is to determine appropriate charges, such as {\em theft, seizing} or {\em robbery}, for criminal cases by analyzing the textual fact descriptions. Automating charge prediction by using NLP technology could significantly reduce the human labor in organizing legal documents, and could be practically useful for an online legal assistant system. 

Existing methods formulate the charge prediction task as a text classification problem, targeting at learning the representation of fact descriptions for prediction. 
Conventional methods~\cite{liu2005classifying,liu2006exploring,lin2012exploiting,sulea2017exploring} design shallow text features to represent fact descriptions. Recently, deep learning provides end-to-end models to learn fact representations from fact descriptions ~\cite{luo2017learning,hu2018few,zhong2018legal}, which achieves the state-of-art result.

In practice, the fact description in a criminal case is written by prosecutors, lawyers, or defendants to state the detail of the criminal case. It comprises a substantial amount of diverse non-normative use of language. 
For example, the cases of robbery in Figure 1 all involve "theft", but the legal term ``theft'' may be implicitly expressed like {\em "stole an electric vehicle"} or {\em "came forward to ride away Ke's white Merida bicycle"}.
Consequently, the representation of fact descriptions may exhibit considerable intra-class variation which may lead to prediction failure at the test stage.
This could be more pronounced for charge classes with only a few examples since the samples are not sufficient for learning a predictive model robust to expression variation. 

To address this issue, we introduce the charge definitions from criminal law to create more robust fact representations for charge prediction. 
We propose to create auxiliary fact representations from the charge definitions to augment the fact representation. 
Those auxiliary representations are essentially projections of the fact description in the semantic space of charge definitions. Our motivation is that the expressions in a fact description should have corresponding formal terms in charge definitions, and those formal terms can provide an alternative view of the expressions in fact description. Note that many of those formal terms are shared across charge classes and are less diverse. Thus, using elements in charge definitions to re-interpret fact description and generate auxiliary representations could have the potential to account for the diversity in the fact description.


\begin{figure}[t]
	\centering
	\includegraphics[]{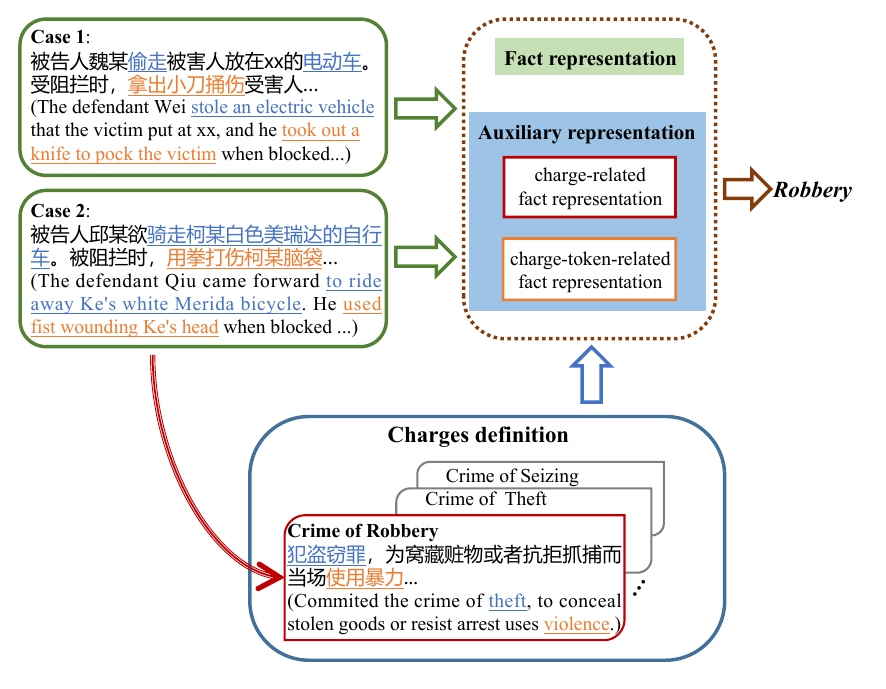}
	\caption{Illustration of our method. The related charges are identified (indicated by the red arrow) via sentence-level attention and aggregated to create the first auxiliary representation, charge-related fact representation. Then key words in cases align to terms in identified charge definitions via word-level attention (aligned words are labeled by the same color), which are then formed as the second auxiliary representation, charge-token-related fact representation.}
	\label{fig1}
\end{figure}

Specifically, we design an integrated sentence- and word-level interaction model to generate two auxiliary fact representations.
We identify the relevant charge definitions through sentence-level interaction between fact description and charge definitions, and then aggregate the holistic features of relevant charge definitions to create the first auxiliary representation, named as charge-related fact representation. The relevant charge definitions identified in the course of producing the first auxiliary representation will also serve for creating the second auxiliary representation. To create the second representation, we further consider finer-grained word-level interaction between the fact description and identified related charge definitions. Relevant words from relevant charge definitions are attended and aggregated through a recurrent neural network to generate the second auxiliary representation, named as charge-token-related fact representation. 
We illustrate our model by an example in Figure 1. Case 1 and case 2 in Figure 1 belong to the same charge class, {\em robbery}, but with different description expressions. With the proposed method, they will be firstly related to the charge definition of {\em robbery}. Then the statements of {\em "stole an electric vehicle"} and {\em "took out a knife to poke the victim"} in case 1, {\em "came forward to ride away Ke's white Merida bicycle"} and {\em "used fist wounding Ke's head"} in case 2 will be softly aligned to the terms "theft" and "use violence" in {\em robbery} definition through attention. By reinterpreting the fact descriptions through aligned terms, those two cases become more similar. 
The final charge prediction is based on the concatenation of the original and auxiliary fact representations, and one can expect the prediction made on this fact representation will be more robust. 

To investigate the advantage of our method on charge prediction, we conduct experiments on two datasets, which consist of criminal cases extracted from the Chinese Judgement web. Experimental results show that our model achieves significant improvement over baselines, especially on classes with few samples. We also conduct ablation studies to analyze the effectiveness of each component in our model, and visualize the impact of introducing charge definitions.



\section{Related Works}
\subsubsection{Charge Prediction} Charge prediction has been studied for years, with the focus on learning representation of fact descriptions in criminal cases and fed into classifiers to make the judgment. At the early stage, ~\cite{liu2005classifying,liu2006exploring,lin2012exploiting,sulea2017exploring} attempt to extract shallow text features from fact descriptions or create hand-crafted features to represent fact descriptions, which are hard to generalize to large datasets due to the diverse expression of fact descriptions. Inspired by the success of deep learning, ~\cite{luo2017learning,ye2018interpretable,hu2018few,zhong2018legal} employ neural models with external information to capture the high-level semantic information. \citeauthor{zhong2018legal} propose the LJP method, modeling multiple legal subtasks as a Directed Acyclic Graph(DAG) and using multi-task learning to assist prediction. 
Further, \citeauthor{luo2017learning} use a separate two-stage scheme to extract the related articles and then attend them attentively to fact representation for charge prediction. \citeauthor{ye2018interpretable} design 10 legal attributes to help the few-shot charges prediction. However, existing charge prediction models all need a large amount of feature engineering, either design features or build relations between subtasks. Instead, we augment fact representation to assist charge prediction by creating auxiliary representation from charge definitions in an end-to-end fashion.

\subsubsection{Attention and Memory} Our model is also related to attention and memory in deep learning ~\cite{bahdanau2014neural,vaswani2017attention,sinha2018hierarchical,weston2014memory,wang2018target,ebesu2018collaborative}. Although researchers propose various neural architectures with memory and attention for NLP problems~\cite{kumar2016ask,wang2017gated,gao2019hybrid}, they either only consider sentence-level or only word-level alignment between sentences. In contrast, we combine them jointly to form auxiliary representation, where sentence-level interaction identifies relevant charges, and a finer-grained word-level interaction on the top of identified charge definitions makes the generated fact representation more robust.

\begin{figure*}[t]
	\centering
	\includegraphics[]{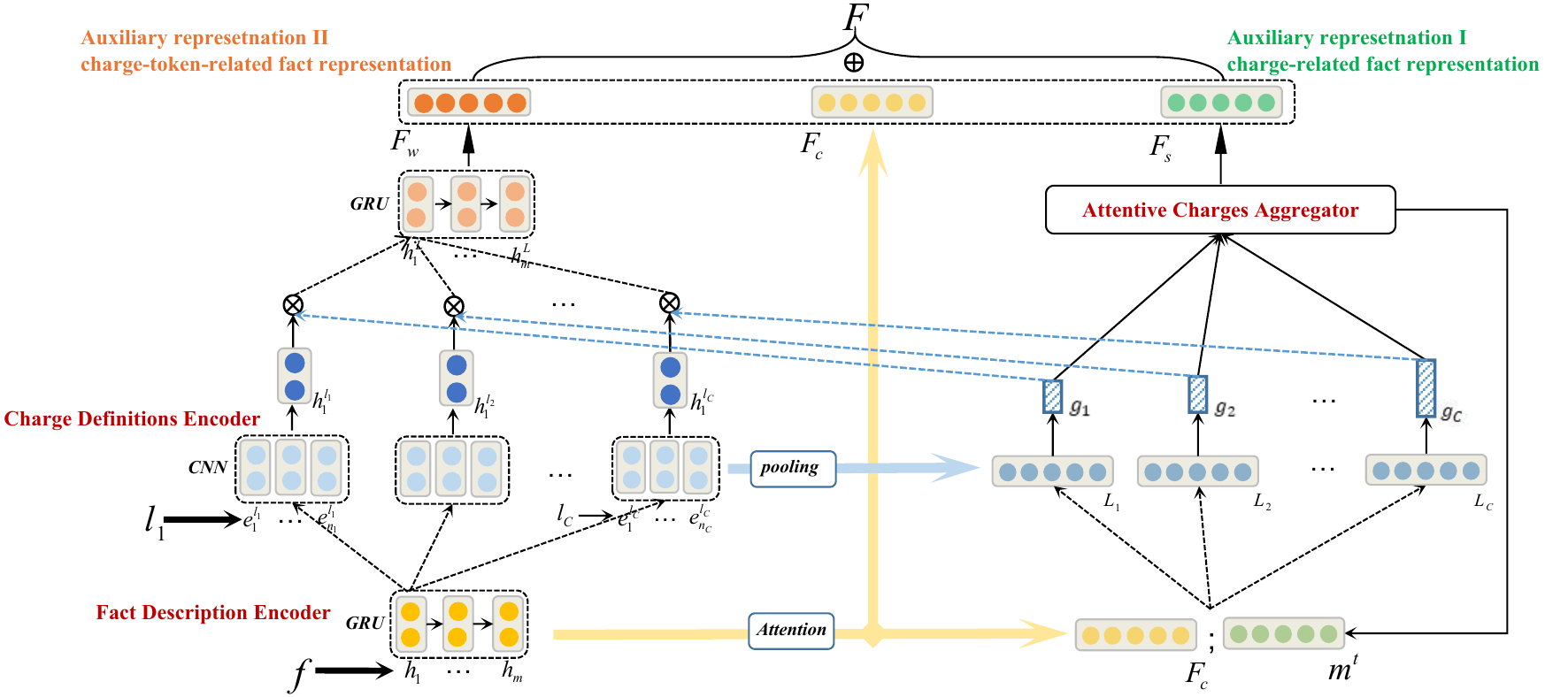}
	\caption{The architecture of our models. Fact description encoder embeds the fact description into the original fact representation $F_{c}$. 
	The right part shows the creation of the first auxiliary representation $F_{s}$: an attentive charge aggregator is iteratively to identify related charges which are then aggregated to generate $F_{s}$.
	The left part shows the creation of the second auxiliary representation $F_{w}$: On top of identified charge definitions, each word in a fact description is represented by the combination of the terms in related charge definitions. The combined intermediate representations are aggregated through a GRU to generate $F_{w}$.
	At last, $F_{c}$, $F_{s}$ and $F_{w}$ are concatenated to form final fact representation $F$.}
	\label{fig2}
\end{figure*}

\section{The proposed model}

\subsection{Problem Formulation}

Charge prediction is to predict the corresponding charges $l$ for a given fact description $f$, where fact description $f$ consists of a sequence of words $\{w_{1}^{f},w_{2}^{f},\cdots,w_{m}^{f}\}$, and its label is a $C$ dimensional multi-hot vector -- a fact description may correspond to multiple charges in $C$ charges. The charge definition for the $i$-th charge $l_{i}$ can be represented as a sequence of words  $\{w_{1}^{l_{i}},w_{2}^{l_{i}},\cdots,w_{n_i}^{l_{i}}\}$.

\subsection{Framework}

To generate a robust fact representation for prediction, we propose an integrated sentence- and word-level interaction model. The architecture of our model is shown in Figure 2. As seen, the final fact representation is the concatenation of three representations.
\begin{itemize}
  \item The original fact representation ($F_{c}$), derived from fact description only and obtained by the fact description encoder. 
  \item The auxiliary representation I, charge-related fact representation ($F_{s}$), aggregated by the holistic representation of related charge definitions that are identified via sentence-level interaction between fact and charge definitions.
  
  \item The auxiliary representation II, charge-token-related fact representation ($F_{w}$), on the top of the identified charge definitions, created by finer-grained word-level interaction between fact and identified charge definitions.
  
\end{itemize}



\subsection{Fact Description Encoder}

Giving an fact description represented by a sequence of word embeddings $f=\{w_{1}^{f},\cdots,w_{i}^{f},\cdots,w_{m}^{f}\}$, we use Gated Recurrent Unite~\cite{cho2014learning} to create a sequence of hidden states for encoding contextual information of each word. 
\begin{align}
    {h}_{i}={GRU}(w_{i}^{f},{h}_{i-1}),
\end{align}
\noindent where ${h}_{i}$ is the hidden state of the GRU at time step $i$. The variable sequence is denoted as $H=\{h_{1},\cdots,h_{i},\cdots,h_{m}\}$.

For a fact description, the words and consequently those hidden variables do not contribute equally to convey the semantic meaning of a fact, and long fact description will involve many less informative words. To suppressing the negative impact of the non-informative words, we use attention mechanism to assign each hidden state in $H$ an importance weight $\alpha_i$.
\begin{align}
    a_{i}=W_{2}\mathit{tanh}(W_{1}{h_{i}}^T),  \\
    \alpha_{i}=\frac{\mathrm{\mathit{exp}(\mathit{a}_{i})} }{\mathrm{\sum_{k=1}^{m}{\mathit{exp}(\mathit{a}_{k})}}},
\end{align}

\noindent where $W_{1}$ and $W_{2}$ are trainable parameters. 
The holistic representation of original fact description ${F_{c}}$ is computed as a weighted sum of $H$:
\begin{align}
    F_{c}=\sum_{i=1}^{m}{\alpha_{i}h_{i}}.
\end{align}

\subsection{Charge Definitions Encoder}

Each charge class is associated with a charge definition, that is, $l_{i}=\{w_{1}^{l_{i}},w_{2}^{l_{i}},\cdots,w_{n_i}^{l_{i}}\}$. We use a CNN to encode the sequence of $n$ words into a sequence of vectors. Since we will deal with a large number of charge definitions, using CNNs~\cite{kim2014convolutional} gives us better training efficiency.
\begin{align}
    e_{j}^{l_{i}}=\mathbf{CNN}(w_{j-\frac{s-1}{2}}^{l_{i}},\cdots,w_{j+\frac{s-1}{2}}^{l_{i}}),
\end{align}

\noindent where the window size of CNN is $s$. Then we sum up these vectors to create the holistic representation of each charge definition.
\begin{align}
    L_{i}=\sum_{j=1}^{n_i}{e_{j}^{l_{i}}}.
\end{align}

We also tried using GRUs to encode $l_{i}$, but they require more computational resources and lead to worse performance. Thus we choose CNN as charge definitions encoder. 

\subsection{Two Auxiliary Fact Representations from Charge Definitions}

The first auxiliary fact representation is created through the sentence-level interaction between the fact description and charge definitions. Its creation process iterates between two steps: identifying related charges and attentively aggregating the holistic representation of charge definitions. After those iterations, relatedness weights of each charge will be obtained and they will also be used as the basis for creating the second auxiliary fact representation. The second auxiliary fact representation is generated from word-level interaction between fact description and identified charge definitions. It uses word-level attention to identify terms that align with the expression in the fact description, and aggregates those terms through a recurrent neural network.

We elaborate the creation of those two auxiliary representations as follows. 

\noindent{\bf Auxiliary Representation I: charge-related fact representation created via sentence-level interaction} 

\subsubsection{Charges Identification} Identifying related charges is realized by calculating an attention weight for each charge to indicate the relatedness. Specifically, we exploit episodic memory attention mechanism~\cite{xiong2016dynamic} to iteratively calculate the attention weight from the correlation between the charge definitions and fact description and memory $m_t$, where $m_t$ can be seen as the summary of already identified charges up to the $t$-th iteration and will be updated at each iteration. With more iterations, the unrelated charges can be filtered out. The memory $m_t$ is initialized with original holistic representation of fact description, that is, $m_0=F_{c}$.

Formally, we use following formulas to calculate the attention weight $g$ of each charge definition at the t-th iteration.
\begin{align}
    z_{i}=[L_{i}\circ F_{c};L_{i}\circ m_{t};|L_{i}-F_{c}|;|L_{i}-m_{t}|], \\
   A_{i}(L_{i},F_{c},m_{t})=W_{2}^{a}\mathit{tanh}(W_{1}^{a}z_{i}), \\
   g_{i}(t)=\frac{\mathrm{\mathit{exp}(A_{i})} }{\mathrm{\sum_{k=1}^{C}{\mathit{exp}(A_{k})}}},
\end{align}

\noindent where $\circ$ is the element-wise product, $|.|$ is the element-wise absolute value, and $;$ represents concatenation of the vectors. $W_{1}^a$ and $W_{2}^a$ are trainable weight matrices.

\subsubsection{Attentive Charges Aggregator} 
Once the attention weight of each charge is calculated, we update the memory by performing weighted summation over charge definition representations. 
\begin{align}
    m_{t+1}=\sum_{i=1}^{C}{g_{i}(t)L_{i}}.
\end{align}

Finally, we concatenate original fact representation with the last memory and the previous memory, and feed them into a fully-connected layer to create the auxiliary charge-related fact representation by using the following equation:
\begin{align}
    F_{s}=fc([F_{c};m_{T};m_{T-1}]),
\end{align}where $fc$ denotes the fully connected layer. 

\noindent{\bf Auxiliary Representation II: charge-token-related fact representation created via word-level interaction} 

In the course of creating the above representation, both fact description and charge definitions are represented by holistic feature vectors. In other words, the interaction between fact and charge definitions is only at the sentence level. The second auxiliary representation steps further to introduce interaction at the word level. Specifically, 
for each hidden variable $h_{k}$ in the fact description, we first compute its matching score towards each $e_{j}^{l_i}$ in each charge definition $l_{i}$ by inner-product. Then $e_{j}^{l_i}$ is attentively aggregated to an intermediate representation $h_{k}^{l_{i}}$:
\begin{align}
    M_j=h_{k}\cdot {e_{j}^{l_i}}^{T}, \\
    \beta_{j}=\frac{\mathrm{\mathit{exp}(M_{j})}}{\mathrm{\sum_{p=1}^{n_i}{\mathit{exp}(M_{p})}}}, \\
   h_{k}^{l_{i}}=\sum_{j=1}^{n_i}{\beta_{j}e_{j}^{l_i}}.
\end{align}

The above intermediate representation is defined w.r.t to each charge definition ${l_{i}}$. In our method, we further perform a weighted summation over $h_{k}^{l_{i}}$ for different charge definition $l_{i}$. The weight is the attention weight $g_{i}(T)$ calculated at the last iteration $T$ in Eq. (9). Using this weight fits our intuition that the terms in the related charges are more relevant to the expressions in the fact description. Formally, we obtain
\begin{align}
    h_{k}^{L}=\sum_{i=1}^{C}g_{i}(T)h_{k}^{l_{i}}.
\end{align}
Note that $h_{k}^{L}$ can be viewed as a projection of $h_k$ in the space spanned by $e_j^{l_i}$.

After obtaining $h_{k}^{L}$ for each word in the fact description, we process the sequence $\{h_1^L,\cdots,h_k^L, \cdots, h_m^L \}$ by a new $GRU$ and obtain the last hidden state $\bar{h}_{l}$:
\begin{align}
    \bar{h}_{t}=GRU(h_{t}^{L},\bar{h}_{t-1}).
\end{align}

We concatenate original and the projected fact representation, and feed them into a fully-connected layer to generate the auxiliary charge-token-related fact representation.
\begin{align}
    F_{w}=fc([F_{c};\bar{h}_{l}]).
\end{align}

\subsection{The Output}

Finally, we concatenate all the generated representations and feed them  into a fully-connected layer to generate the final fact representation $F$.
\begin{align}
    F=fc([F_{c};F_{s};F_{w}]).
\end{align}
$F$ is then passed to the classifier layer to make charge prediction.

The loss function for training is as follows:
\begin{align}
    Loss = -\sum_{i=1}^{N}{\sum_{l=1}^{C}{y_{il}\mathit{log}(o_{il})}},
\end{align}
\noindent where $N$ is the number of training data, $C$ is the number of charges. $y_{il} \in \{0,1\}$ and $o_{il}$ is the estimated likelihood of the $l$-th charge being true. 

\section{Experiments}

In order to verify the effectiveness of our model on criminal charges prediction, we conduct experiments on two real-world datasets with different scales to compare our model against several baselines. Further analyses are also made to validate the significance of introducing charge definitions and various components of our model.

\subsection{Setup}

\begin{table}[]
	\caption{Statistics of datasets.}
	\begin{center}
\begin{tabular}{lcc}
\hline
\multicolumn{1}{l|}{\textbf{Datasets}} & \multicolumn{1}{l|}{\textbf{CAIL150K}} & \textbf{CAIL30K}  \\ \hline
\multicolumn{1}{l|}{Traning samples} & \multicolumn{1}{c|}{154592} & \multicolumn{1}{c}{32506} \\ 
\multicolumn{1}{l|}{Test samples}    & \multicolumn{1}{c|}{32500}  & \multicolumn{1}{c}{32500} \\ 
\multicolumn{1}{l|}{Charge classes}  & \multicolumn{1}{c|}{202}    & \multicolumn{1}{c}{168}   \\ 
\hline
\end{tabular}
	\end{center}
	\label{table1} 
\end{table}

\subsubsection{Dataset} We use publicly available datasets from ~\cite{xiao2018cail2018} to conduct our experiments. There are two datasets with different scales: \verb|CAIL150K| dataset and \verb|CAIL30K| dataset. The criminal cases in these datasets are collected from the China Judgment Online\footnote{http://wenshu.court.gov.cn/} with a single defendant. Table 1 shows the descriptive statistics of used datasets\footnote{The training sets of CAIL150K and CAIL30K are exercise\_contest/data\_train.json and final\_contest.json separately in CAIL2018 file.}. It is worth noting that in these two datasets the distribution of charges is quite imbalanced. In \verb|CAIL150K|, the top 30 most frequent charges cover 60\% cases, and the 31\% charges in the training set have less than 100 cases, taking up only 1.88\% of the total number of cases. \verb|CAIL30K| is a smaller dataset. In its training set, 42\% charges have less than 10 cases, taking up only 0.89\% of the total number of cases. The small number of samples makes it challenging to train a model that performs well on low-frequency classes.

As for charge definitions, they are extracted from articles in the Criminal Law of the People's Republic of China. Specifically, in criminal law, except for articles irrelevant to specific charges, each article may include more than one charges, their corresponding charge definitions, and punishment. We use regular expressions to extract charge definitions, and merge charge definitions scattered in multiple articles. A snippet of cases and charge definitions is illustrated in Figure 1.

\subsubsection{Training setup} As all the sentences in charge definitions and fact descriptions are written in Chinese without word segmenting, we apply jieba\footnote{https://github.com/fxsjy/jieba} for word cut. We set the maximum length of fact description to 500, charge definitions to 110. We use pre-trained GloVe~\cite{dong2014adaptive} vectors as our initial word embeddings. In practice, we choose the 64 dimensional embedding vectors trained on {\verb|baidubaike|}. The iteration time $T$ in Eq. (9) is set as 3. Adam~\cite{kingma2014adam} is used as the optimizer and the learning rate is initialized as 0.005 and halved in every other epoch.

\begin{table*}[t]
	\caption{The experimental results [\%] of baselines and our model on two datasets. Four different types of models are separated by lines and the best scores are highlight in bold font.}
	\begin{center}
\begin{tabular}{ll|llll|llll}
\hline
\multicolumn{2}{c}{\textbf{Datasets}}  & \multicolumn{4}{c}{\textbf{CAIL150K}} & \multicolumn{4}{c}{\textbf{CAIL30K}} \\
\multicolumn{2}{c}{\textbf{Model}} & \multicolumn{1}{c}{\textbf{Acc.}} & \multicolumn{1}{c}{\textbf{MP}} & \multicolumn{1}{c}{\textbf{MR}} & \multicolumn{1}{c}{\textbf{MF1}} & \multicolumn{1}{c}{\textbf{Acc.}} & \multicolumn{1}{c}{\textbf{MP}} & \multicolumn{1}{c}{\textbf{MR}} & \multicolumn{1}{c}{\textbf{MF1}} \\  \hline
\multirow{4}{*}{Not using charge definitions} 
& \textbf{TFIDF+SVM}        & 71.87     & 79.71   & 56.84    & 63.32    & 49.13  & 31.48   & 19.98    & 22.06  \\
& \textbf{CNN\_classify}    & 79.23     & 70.80   & 62.27    & 64.97    & 52.75  & 23.64   & 21.95    & 20.59  \\
& \textbf{GRU\_classify}    & 77.33     & 72.45   & 57.42    & 61.54    & 56.14  & 23.99   & 22.81    & 21.51  \\
& \textbf{HLSTM\_classify}  & 73.15     & 51.45   & 43.82    & 46.06    & 25.34  & 7.69    & 6.34     & 6.15   \\
\hline
\multirow{1}{*}{Using multiple tasks} 
& \textbf{LJP}           & 25.26    & 25.78    & 24.32    & 25.55   & 15.29   & 15.45  & 15.68   & 15.56   \\ \hline
\multirow{2}{*}{Match with charge definitions}    
& \textbf{TFIDF match}      & 13.03    & 31.21    & 40.29    & 26.52   & 12.19   & 37.37  & 35.41   & 27.60    \\
& \textbf{Siamese CNN}      & 72.98    & 74.52    & 64.64    & 66.55   & 50.66   & 32.74  & 33.74   & 29.28    \\
\hline
\multirow{4}{*}{Augment with charge definitions} 
& \textbf{Fact-Law AN}   & 75.61    & 58.89    & 52.30    & 53.62   & 60.73   & 28.15  & 25.16   & 24.79  \\
& \textbf{GA\_Reader}     & 73.78    & 74.68    & 66.59    & 68.21   & 54.95   & 39.29  & 34.05   & 33.03   \\
& \textbf{MemNet}        & 80.18    & 80.09    & 67.13    & 70.78   & 62.40   & 32.62  & 27.54   & 27.64   \\
& \textbf{Ours}     &      \textbf{81.05}    & \textbf{82.06}      & \textbf{68.33}  & \textbf{72.43}   & \textbf{67.99}   & \textbf{46.13}   & \textbf{36.00}   & \textbf{37.62} \\
\hline
\end{tabular}
	\end{center}
	\label{tabel2} 
\end{table*}

\subsubsection{Baselines} We compare our model against several text classification models and charge prediction
methods, which can be categorized into four categories:

(1){\bf Not using charge definitions for classification.} We employ TFIDF~\cite{salton1988term} to extract text features from fact descriptions and use linear SVMs~\cite{suykens1999least} for charge prediction ({\bf TFIDF+SVM}). We also implement deep learning models, such as multi-layers Convolution Neural Network(CNN) ~\cite{kim2014convolutional} ({\bf CNN\_classify}), Gated Recurrent Unite (GRU)~\cite{cho2014learning} ({\bf GRU\_classify}) and hierarchical LSTM ~\cite{sinha2018hierarchical} ({\bf HLSTM\_classify}) for fact descriptions encoding and classification. 

(2){\bf Using multi-task learning for classification.} Our method is somehow related to {\bf LJP}~\cite{zhong2018legal}, which introduces related legal tasks and use multi-task learning to train a better fact representation. We also re-implement it to compare with our method.

(3){\bf Matching the fact description with charge definitions for classification.} We exploit TFIDF to extract text features from fact descriptions and charge definitions, then compare the fact description with each charge definition ({\bf TFIDF match}) to find the best matched charges. We also train a Siamese CNN~\cite{koch2015siamese} ({\bf Siamese CNN}) to match the representations of fact description and charge definitions.

(4){\bf Augmenting fact description with charge definitions for classification.} We implement {\bf Fact-Law AN} model that \citeauthor{luo2017learning} propose to use relevant law articles, selected by SVMs, to serve as a legal basis for encoding the fact description. To demonstrate the advantage of our model in considering sentence- and word-level interaction jointly, we also implement improved memory network~\cite{kumar2016ask} ({\bf MemNet}) and {\bf GA\_Reader}~\cite{wang2017gated}. These two methods are designed for question-answer task, which employ multi-iterative interaction between query and document at sentence- and word-level respectively for answer prediction. In the implementation, we replace query and document in {\bf GA\_Reader} and {\bf MemNet} with the fact description and charge definitions.

\begin{table}[t]
	\caption{The experimental results of ablation test of our model on CAIL150K dataset.}
	\begin{center}
	\begin{tabular}{p{2.6cm}p{0.9cm}p{0.9cm}p{0.9cm}p{0.9cm}}
		\hline 
		{\textbf{Models}} & { \textbf{Acc.}} & { \textbf{MP}} & { \textbf{MR}} & { \textbf{MF1}}  \\ \hline 
		{\em \textbf{Ours}}       & \textbf{81.05}    & \textbf{82.06}      & \textbf{68.33}  & \textbf{72.43}   \\
		{\em  \hspace*{1.5em}   w/o  Fc}  & 80.31  &79.12  &66.88  &70.55  \\
		{\em  \hspace*{1.5em}   w/o  Fs,Fw}  & 77.33  & 72.45  & 57.42  & 61.54   \\
		{\em  \hspace*{1.5em}   w/o  Fw}     & 79.50  & 78.86  & 66.18  & 69.86   \\
		{\em  \hspace*{1.5em}   w/o  Fs}     & 80.62  & 80.54  & 66.97  & 71.28   \\
		{\em  \hspace*{1.5em}   w/o  Fs,$g_i$}  & 80.54  & 76.90  & 64.34  & 67.98   \\
		\hline
	\end{tabular}
	\end{center}
	\label{table6}
\end{table}

\subsection{Results}

\subsubsection{Evaluation Metrics} We employ accuracy (Acc.), macro-precision (MP), macro-recall (MR) and macro-F1 (MF1) as our evaluation metrics. The macro-precision/recall/F1 are calculated by averaging the precision, recall and F1 of each charge, which are metrics commonly used for multilabel classification task. 

\subsubsection{Overall Evaluation Results} Experimental results on two scale datasets are shown in Table 2. The observations are as followings:
\begin{itemize}
  \item Generally speaking, models without incorporating charge definitions ({\bf TFIDF+SVM}, {\bf CNN\_classify}, {\bf GRU\_classify} and {\bf HLSTM\_classify}) perform inferior to their charge-definition-incorporated counterparts. This is evident by their lower MF1 scores (MF1 is a more comprehensive score for evaluating multi-label classification than Acc., MP, and MR). This observation clearly demonstrates the benefit of introducing charge definitions to assist charge prediction.

  \item Incorporating charge definitions through matching based approaches ({\bf TFIDF match} and {\bf Siamese CNN}) works to some extent, although their performance is still worse than methods using more sophisticated interaction between fact description and charge definitions, i.e. {\bf GA\_Reader}, {\bf MemNet} and {\bf Ours}.

  \item Methods that Augment fact representation with charge definitions through end-to-end schema ({\bf GA\_Reader}, {\bf MemNet} and {\bf Ours}) attain better results than {\bf Fact-Law AN}.
  The latter uses a separated two-stage framework to first identify the related charge definitions. This observation shows the importance of the end-to-end design. In addition, compared with {\bf GA\_Reader} and {\bf MemNet}, which performs either sentence- or word-level interaction, our approach achieves better performance through considering sentence- and word-level interaction jointly.
  
   \item Our proposed model outperforms other baselines on two datasets. The improvement is especially significant on the \verb|CAIL30K| dataset: our method surpasses the second best about 5\% in MF1. Since the \verb|CAIL30K| contains more classes with few training samples, the excellent performance of our approach suggests that our auxiliary representations may help to improve the generalization performance for classes with few samples.

  \item Finally, we compare our method against {\bf LJP}. Like our method, {\bf LJP} also uses external information for building the fact representation. Different from our method, they introduce multiple related tasks and adopt the multi-task learning for representation training. As shown in Table 2, we can see that our method achieves superior performance than {\bf LJP}.   
  
\end{itemize}

\subsection{Further Analysis}

\begin{figure}[]
	\includegraphics[]{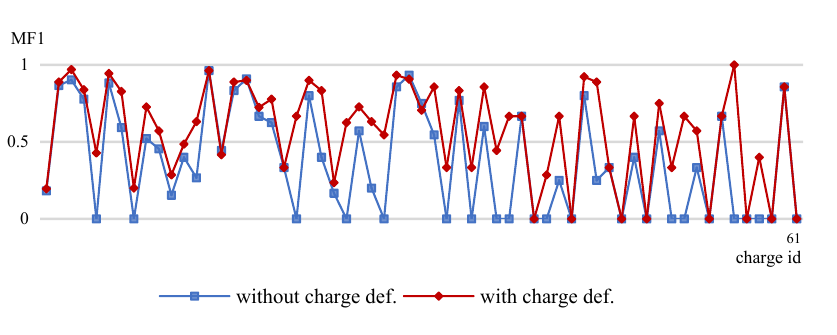}
	\caption{Results of the impact of exploiting charge definitions for charges predicting under the MF1 metric. The charge ids are those classes with training samples less than 100 in CAIL150K dataset.}
	\label{fig3}
\end{figure}

\subsubsection{Ablation Test} We conduct ablation studies to verify the effectiveness of various components in our method. We consider several variations of our approach by removing some components of our model. The result is shown in Table 3. As seen, only using fact descriptions without any level auxiliary fact representations ({\em w/o Fs,Fw}) yields the worst performance, which proves the importance of the use of charge definitions. After adding either the sentence-level ({\em w/o Fw}) or the word-level auxiliary fact representation ({\em w/o Fs}), the performance can be significantly improved. It is observed that the performance of only adding charge-token-related fact representation ({\em w/o Fs}) is better than only adding charge-related fact representation ({\em w/o Fw}). 
We also created a variant of our method without using attention weight $g_{i}$ of each charge from Eq. (9) in the process of generating charge-token-related fact representation ({\em w/o Fs,$g_i$}), which is implemented by setting the attention weight $g_i$ to $\frac{1}{C}$ instead of generated from charge identification part. It can be observed that the performance of {\em w/o Fs,$g_i$} declines. This suggests that the two-level attention is necessary and using them jointly can get the best performance.



\subsubsection{Impact of Exploiting charge definitions} We analyze the effects of incorporating generated auxiliary fact representations for classes with few training data. As shown in Figure 3, we study the results of classes with less than 100 samples on \verb|CAIL150K| dataset. We can find that the MF1 measure of many charges is zero if auxiliary representations are not used, and the results can be improved significantly if we add auxiliary representation from charge definitions. This observation highlights the benefit of introducing auxiliary representations for handling small sample cases.

\begin{figure}[]
	\includegraphics[]{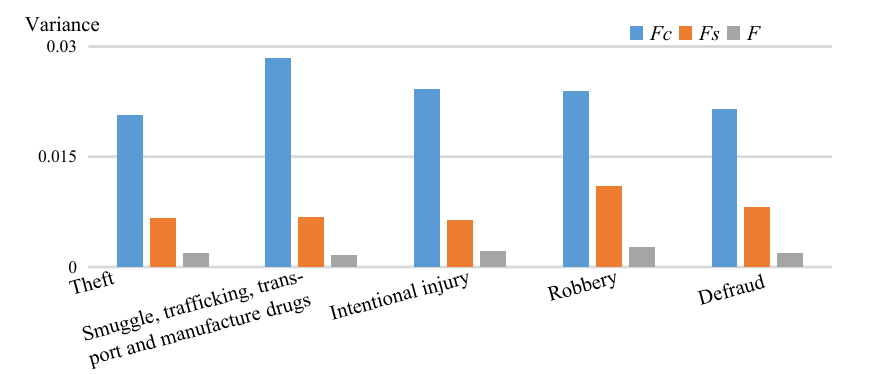}
	\caption{Intra-class variance of different fact representations of the top-5 frequent classes in CAIL150K dataset. $Fc$ is fact representation only learned from fact description, $Fs$ is the $F_{c}$ augmented with charge-related fact representation, and $F$ is the $F_{c}$ augmented with all auxiliary fact representations.}
	\label{figure4}
\end{figure}

\subsubsection{Intra-class variance of different fact representations} To investigate whether the fact representation of our method is more stable, we conduct the following experiment: we calculate the variance along each dimension of fact representations from five classes with the most amount of samples, and then use the average variance along all dimensions as an indicator of the intra-class variance of different fact representations. 
As shown in Figure 4, fact representation ($Fc$) only learned from fact description yields the largest intra-class variance. After augmenting fact representation from charge definitions through sentence-level interaction ($Fs$), the intra-class variance declines greatly. Specially, the final fact representation ($F$) with two auxiliary representations incorporated attains an even lower intra-class variance.

\subsubsection{Case study} Finally, we select a representative robbery case to give an intuitive illustration of the attention results on the sentence- and word-level interaction. As shown in Table 4, the case describes that the defendant is convicted of robbery due to stealing property and poking the victim to resist arrest. On the sentence-level interaction, with the increasing of iteration in Eq. (9), our model narrows down the candidate charges and finally identifies the correct related charges. We choose the iteration times as 3 since the performance cannot improve with more iterations.

On the word-level interaction, the attention mechanism makes the words in fact description align with the formal terms in charge definitions. To demonstrate this mechanism, Figure 5 shows for the words in fact description, which terms are focused on in the charge definition of robbery. The identified keywords in fact description are "electric vehicle", "resisting arrest" and "a knife", which correspond to key terms in robbery definition--"stolen goods", "resist arrest" and "use violence".

\begin{CJK}{UTF8}{gbsn}
\begin{table}[t]
	\caption{Attention map of sentence-level attention of robbery case. t1, t2, and t3 represent the iteration times in Eq. (9). The color darker means the charges are more related to the fact.}
	\begin{center}
\begin{tabular}{l|lll}
\hline
\multicolumn{4}{l}{\begin{tabular}[c]{@{}l@{}}\footnotesize {\bf Fact description:} 被告人偷盗电动车，被受害人阻拦时，\\
The Defendant stole an electric vehicle, when blocked,  \\\footnotesize 为抗拒抓捕，拿出一把小刀捅伤受害人... \\ he took out a knife to pock the victim to resist arrest...\\\footnotesize {\bf Charge: Robbery}\end{tabular}} \\ \hline
\textbf{Top5 Related Charges}               & \multicolumn{1}{l|}{\textit{\textbf{\hspace*{0.3em}t1\hspace*{0.3em}}}}         & \multicolumn{1}{l|}{\textit{\textbf{\hspace*{0.3em}t2\hspace*{0.3em}}}}         & \multicolumn{1}{l}{\textit{\textbf{\hspace*{0.3em}t3\hspace*{0.3em}}}}        \\ \hline
Robbery       & \cellcolor[HTML]{2E75B6}  &   \cellcolor[HTML]{266092}  & \cellcolor[HTML]{010066}  
\\ \cline{1-1}
Intentional injury   & \cellcolor[HTML]{A7C4E6}   & \cellcolor[HTML]{A9D7EE}   & \cellcolor[HTML]{F0FFF0} \\ \cline{1-1}
Theft       & \cellcolor[HTML]{A9D7EE}      & \cellcolor[HTML]{D8E3F3}    &             \\ \cline{1-1}
Negligent act causing severe injury         & \cellcolor[HTML]{DAE3F3}    &       &     \\ \cline{1-1}
Endangering public security                 & \cellcolor[HTML]{E2F0D9}    &             \\ \hline
\end{tabular}
\end{center}
\end{table}
\end{CJK}

\begin{figure}[t]
	\centering 
	\includegraphics[]{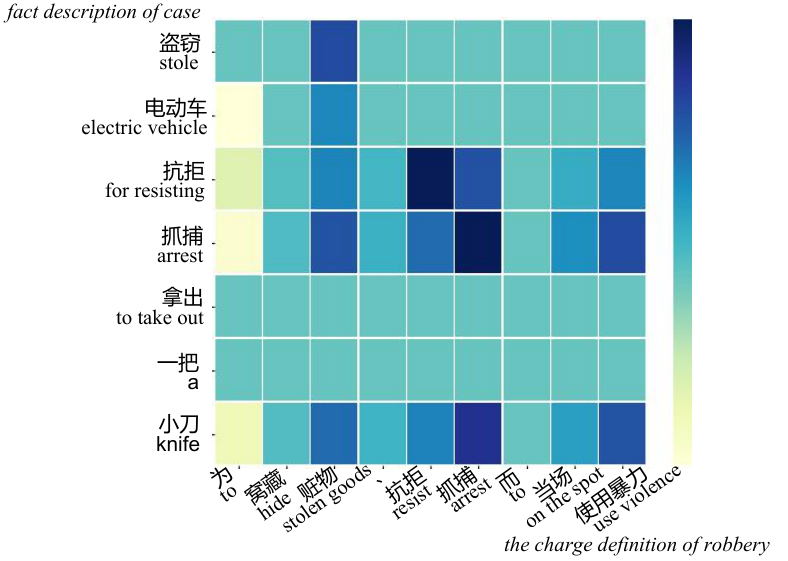}
	\caption{Attention map of word-level attention between robbery case and the charge definition of robbery. The dark color means a large value.}
	\label{figure5}
\end{figure}

\section{Conclusion}

In this work, we focus on the task of multilabel charge prediction for given fact descriptions of criminal cases. To address the problem of having a large expression variance in fact descriptions due to informal language use, we introduce charge definitions from criminal law to create auxiliary representations of the fact descriptions. The experimental results on two datasets show the effectiveness of our model on charge prediction. The significant improvement on the classes with few training data validate that our method can benefit the small sample training scenario and the two-level auxiliary fact representations can help the model to generalize to the unseen description. 


\bibliography{aaai20}

\begin{thebibliography}{}

\bibitem[\protect\citeauthoryear{Bahdanau, Cho, and
  Bengio}{2014}]{bahdanau2014neural}
Bahdanau, D.; Cho, K.; and Bengio, Y.
\newblock 2014.
\newblock Neural machine translation by jointly learning to align and
  translate.
\newblock {\em arXiv preprint arXiv:1409.0473}.

\bibitem[\protect\citeauthoryear{Cho \bgroup et al\mbox.\egroup
  }{2014}]{cho2014learning}
Cho, K.; Van~Merri{\"e}nboer, B.; Gulcehre, C.; Bahdanau, D.; Bougares, F.;
  Schwenk, H.; and Bengio, Y.
\newblock 2014.
\newblock Learning phrase representations using rnn encoder-decoder for
  statistical machine translation.
\newblock {\em arXiv preprint arXiv:1406.1078}.

\bibitem[\protect\citeauthoryear{Dong \bgroup et al\mbox.\egroup
  }{2014}]{dong2014adaptive}
Dong, L.; Wei, F.; Tan, C.; Tang, D.; Zhou, M.; and Xu, K.
\newblock 2014.
\newblock Adaptive recursive neural network for target-dependent twitter
  sentiment classification.
\newblock In {\em Proceedings of the 52nd annual meeting of the association for
  computational linguistics (volume 2: Short papers)}, volume~2,  49--54.

\bibitem[\protect\citeauthoryear{Ebesu, Shen, and
  Fang}{2018}]{ebesu2018collaborative}
Ebesu, T.; Shen, B.; and Fang, Y.
\newblock 2018.
\newblock Collaborative memory network for recommendation systems.
\newblock In {\em The 41st International ACM SIGIR Conference on Research \&
  Development in Information Retrieval},  515--524.
\newblock ACM.

\bibitem[\protect\citeauthoryear{Gao \bgroup et al\mbox.\egroup
  }{2019}]{gao2019hybrid}
Gao, T.; Han, X.; Liu, Z.; and Sun, M.
\newblock 2019.
\newblock Hybrid attention-based prototypical networks for noisy few-shot
  relation classification.

\bibitem[\protect\citeauthoryear{Hu \bgroup et al\mbox.\egroup
  }{2018}]{hu2018few}
Hu, Z.; Li, X.; Tu, C.; Liu, Z.; and Sun, M.
\newblock 2018.
\newblock Few-shot charge prediction with discriminative legal attributes.
\newblock In {\em Proceedings of the 27th International Conference on
  Computational Linguistics},  487--498.

\bibitem[\protect\citeauthoryear{Kim}{2014}]{kim2014convolutional}
Kim, Y.
\newblock 2014.
\newblock Convolutional neural networks for sentence classification.
\newblock {\em arXiv preprint arXiv:1408.5882}.

\bibitem[\protect\citeauthoryear{Kingma and Ba}{2014}]{kingma2014adam}
Kingma, D.~P., and Ba, J.
\newblock 2014.
\newblock Adam: A method for stochastic optimization.
\newblock {\em arXiv preprint arXiv:1412.6980}.

\bibitem[\protect\citeauthoryear{Koch, Zemel, and
  Salakhutdinov}{2015}]{koch2015siamese}
Koch, G.; Zemel, R.; and Salakhutdinov, R.
\newblock 2015.
\newblock Siamese neural networks for one-shot image recognition.
\newblock In {\em ICML deep learning workshop}, volume~2.

\bibitem[\protect\citeauthoryear{Kumar \bgroup et al\mbox.\egroup
  }{2016}]{kumar2016ask}
Kumar, A.; Irsoy, O.; Ondruska, P.; Iyyer, M.; Bradbury, J.; Gulrajani, I.;
  Zhong, V.; Paulus, R.; and Socher, R.
\newblock 2016.
\newblock Ask me anything: Dynamic memory networks for natural language
  processing.
\newblock In {\em International conference on machine learning},  1378--1387.

\bibitem[\protect\citeauthoryear{Lin \bgroup et al\mbox.\egroup
  }{2012}]{lin2012exploiting}
Lin, W.-C.; Kuo, T.-T.; Chang, T.-J.; Yen, C.-A.; Chen, C.-J.; and Lin, S.-d.
\newblock 2012.
\newblock Exploiting machine learning models for chinese legal documents
  labeling, case classification, and sentencing prediction.
\newblock {\em Processdings of ROCLING}  140.

\bibitem[\protect\citeauthoryear{Liu and Hsieh}{2006}]{liu2006exploring}
Liu, C.-L., and Hsieh, C.-D.
\newblock 2006.
\newblock Exploring phrase-based classification of judicial documents for
  criminal charges in chinese.
\newblock In {\em International Symposium on Methodologies for Intelligent
  Systems},  681--690.
\newblock Springer.

\bibitem[\protect\citeauthoryear{Liu and Liao}{2005}]{liu2005classifying}
Liu, C.-L., and Liao, T.-M.
\newblock 2005.
\newblock Classifying criminal charges in chinese for web-based legal services.
\newblock In {\em Asia-Pacific Web Conference},  64--75.
\newblock Springer.

\bibitem[\protect\citeauthoryear{Luo \bgroup et al\mbox.\egroup
  }{2017}]{luo2017learning}
Luo, B.; Feng, Y.; Xu, J.; Zhang, X.; and Zhao, D.
\newblock 2017.
\newblock Learning to predict charges for criminal cases with legal basis.
\newblock {\em arXiv preprint arXiv:1707.09168}.

\bibitem[\protect\citeauthoryear{Salton and Buckley}{1988}]{salton1988term}
Salton, G., and Buckley, C.
\newblock 1988.
\newblock Term-weighting approaches in automatic text retrieval.
\newblock {\em Information processing \& management} 24(5):513--523.

\bibitem[\protect\citeauthoryear{Sinha \bgroup et al\mbox.\egroup
  }{2018}]{sinha2018hierarchical}
Sinha, K.; Dong, Y.; Cheung, J. C.~K.; and Ruths, D.
\newblock 2018.
\newblock A hierarchical neural attention-based text classifier.
\newblock In {\em Proceedings of the 2018 Conference on Empirical Methods in
  Natural Language Processing},  817--823.

\bibitem[\protect\citeauthoryear{Sulea \bgroup et al\mbox.\egroup
  }{2017}]{sulea2017exploring}
Sulea, O.-M.; Zampieri, M.; Malmasi, S.; Vela, M.; Dinu, L.~P.; and van
  Genabith, J.
\newblock 2017.
\newblock Exploring the use of text classification in the legal domain.
\newblock {\em arXiv preprint arXiv:1710.09306}.

\bibitem[\protect\citeauthoryear{Suykens and
  Vandewalle}{1999}]{suykens1999least}
Suykens, J.~A., and Vandewalle, J.
\newblock 1999.
\newblock Least squares support vector machine classifiers.
\newblock {\em Neural processing letters} 9(3):293--300.

\bibitem[\protect\citeauthoryear{Vaswani \bgroup et al\mbox.\egroup
  }{2017}]{vaswani2017attention}
Vaswani, A.; Shazeer, N.; Parmar, N.; Uszkoreit, J.; Jones, L.; Gomez, A.~N.;
  Kaiser, {\L}.; and Polosukhin, I.
\newblock 2017.
\newblock Attention is all you need.
\newblock In {\em Advances in neural information processing systems},
  5998--6008.

\bibitem[\protect\citeauthoryear{Wang \bgroup et al\mbox.\egroup
  }{2017}]{wang2017gated}
Wang, W.; Yang, N.; Wei, F.; Chang, B.; and Zhou, M.
\newblock 2017.
\newblock Gated self-matching networks for reading comprehension and question
  answering.
\newblock In {\em Proceedings of the 55th Annual Meeting of the Association for
  Computational Linguistics (Volume 1: Long Papers)},  189--198.

\bibitem[\protect\citeauthoryear{Wang \bgroup et al\mbox.\egroup
  }{2018}]{wang2018target}
Wang, S.; Mazumder, S.; Liu, B.; Zhou, M.; and Chang, Y.
\newblock 2018.
\newblock Target-sensitive memory networks for aspect sentiment classification.
\newblock In {\em Proceedings of the 56th Annual Meeting of the Association for
  Computational Linguistics (Volume 1: Long Papers)},  957--967.

\bibitem[\protect\citeauthoryear{Weston, Chopra, and
  Bordes}{2014}]{weston2014memory}
Weston, J.; Chopra, S.; and Bordes, A.
\newblock 2014.
\newblock Memory networks.
\newblock {\em arXiv preprint arXiv:1410.3916}.

\bibitem[\protect\citeauthoryear{Xiao \bgroup et al\mbox.\egroup
  }{2018}]{xiao2018cail2018}
Xiao, C.; Zhong, H.; Guo, Z.; Tu, C.; Liu, Z.; Sun, M.; Feng, Y.; Han, X.; Hu,
  Z.; Wang, H.; et~al.
\newblock 2018.
\newblock Cail2018: A large-scale legal dataset for judgment prediction.
\newblock {\em arXiv preprint arXiv:1807.02478}.

\bibitem[\protect\citeauthoryear{Xiong, Merity, and
  Socher}{2016}]{xiong2016dynamic}
Xiong, C.; Merity, S.; and Socher, R.
\newblock 2016.
\newblock Dynamic memory networks for visual and textual question answering.
\newblock In {\em International conference on machine learning},  2397--2406.

\bibitem[\protect\citeauthoryear{Ye \bgroup et al\mbox.\egroup
  }{2018}]{ye2018interpretable}
Ye, H.; Jiang, X.; Luo, Z.; and Chao, W.
\newblock 2018.
\newblock Interpretable charge predictions for criminal cases: Learning to
  generate court views from fact descriptions.
\newblock {\em arXiv preprint arXiv:1802.08504}.

\bibitem[\protect\citeauthoryear{Zhong \bgroup et al\mbox.\egroup
  }{2018}]{zhong2018legal}
Zhong, H.; Zhipeng, G.; Tu, C.; Xiao, C.; Liu, Z.; and Sun, M.
\newblock 2018.
\newblock Legal judgment prediction via topological learning.
\newblock In {\em Proceedings of the 2018 Conference on Empirical Methods in
  Natural Language Processing},  3540--3549.

\end{thebibliography}
\bibliographystyle{aaai}


\end{document}